# Siamese Infrared and Visible Light Fusion Network for RGB-T Tracking

Jingchao Peng[1], Haitao Zhao[2], Zhengwei Hu[3], Yi Zhuang[4], and Bofan Wang[5]
East China University of Science and Technology,
Automation Department, School of Information Science and Engineering
{[1]pjc, [3]zwh, [4]y30190773, [5]y30190751}@mail.ecust.edu.cn, [2]htz@ecust.edu.cn

*Abstract*—Due to the different photosensitive properties of infrared and visible light, the registered RGB-T image pairs shot in the same scene exhibit quite different characteristics. This paper proposes a siamese infrared and visible light fusion Network (SiamIVFN) for RBG-T image-based tracking. SiamIVFN contains two main subnetworks: a complementary-feature-fusion network (CFFN) and a contribution-aggregation network (CAN). CFFN utilizes a two-stream multilayer convolutional structure whose filters for each layer are partially coupled to fuse the features extracted from infrared images and visible light images. CFFN is a feature-level fusion network, which can cope with the misalignment of the RGB-T image pairs. Through adaptively calculating the contributions of infrared and visible light features obtained from CFFN, CAN makes the tracker robust under various light conditions. Experiments on two RGB-T tracking benchmark datasets demonstrate that the proposed SiamIVFN has achieved state-of-the-art performance. The tracking speed of SiamIVFN is 147.6FPS, the current fastest RGB-T fusion tracker.

*Index Terms*—Object tracking, deep learning, fusion tracking, siamese network

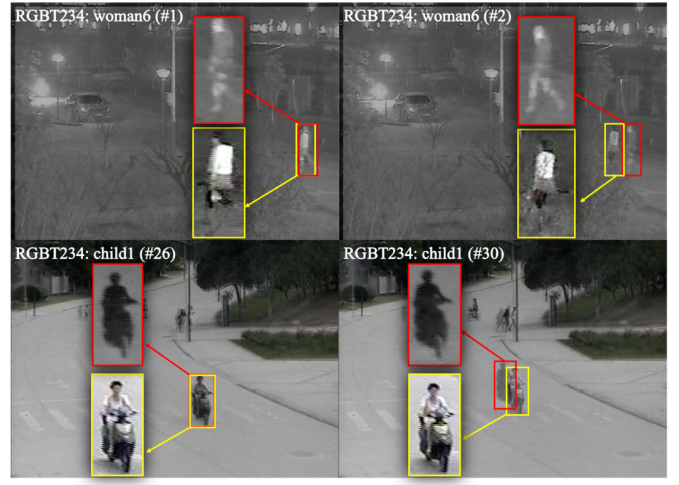

Fig. 1. Superimposed image of infrared and visible light image pair. The weight of the visible light image is 0.6, and the weight of the infrared image is 0.4. The object in the yellow box with rich texture, details, and color is from the visible light image. The object in the red box that can only be roughly distinguished by the silhouette is from the infrared image.

## I. Introduction

OBJECT tracking is an essential task in computer vision and video technology. In the past decades, deep convolutional networks have been successfully applied in different fields, especially in object tracking [2, 18, 22, 34]. Visible light images have rich texture information and high contrast, which is useful for object tracking tasks. However, in weak light level conditions such as cloudy nights or light visibility conditions such as aerosols, visible-light-images-based object tracking may be difficult to function. Unlike visual light images, thermal infrared images mainly recording the thermal radiation of objects are stable under drastic changes in light or weak light level conditions [4, 41]. Infrared images can penetrate rain, fog, and snow. Nevertheless, infrared images lack texture information rich in visible light images and have low contrast. Due to the complementarity between infrared and visible light images, object tracking based on the fusion of infrared and visible light images has attracted more and more attention [11, 13, 14, 15].

The existing fusion tracking based on infrared and visible light images (or so-called RGB-T fusion tracking) methods focuses on supplementing thermal information to assistant visible-light-image-based tracking [9, 26, 28]. Their purpose is to compensate for the visible light image in the deteriorated light conditions. RGB-T fusion can be divided into pixel-level fusion, feature-level fusion, and decision-level fusion [37].

Pixel-level fusion is to fuse the rigorous registered image pairs pixel by pixel and then perform object tracking based on the merged images [37]. Pixel-level fusion is sensitive to noise and has a high demand for image registration [6]. Decision-level fusion performs tracking task respectively on RGB images and thermal images and then aggregate two different tracking results (such as the position and the size of the tracking object) to obtain the final tracking result [36]. That is, separate images are processed individually and then fed into the fusion algorithm. Unlike pixel-level fusion, decision-level fusion does not require obtaining individual pixel values at the very same locations by interpolation. However, decision-level fusion pays little attention to the feature complementarity between infrared images and visible light images, leading to unreliable tracking effects that rely on a single pattern [6].

Feature-level fusion is to extract and fuse features of RGB-T image pairs then utilize the fused features for tracking [24, 27, 38]. In this way, the tracking result may not rely too much on a single pattern tracker or original strict registered image pairs [6, 37]. Although a spatial registration step is still necessary for feature extraction and fusion, feature-level fusion allows for explicitly handling localization uncertainty, for instance, due to the misalignment of the image pairs.

To visually demonstrate the characteristics of the RGB-T pairs, we linearly superimpose the infrared image and the visible light image, as shown in Fig. 1. On the one hand, because two images shot simultaneously at the same place, the image pair should contain common features. On the other hand, since cameras capture two images with different sensor types, the infrared image and the visible light image have individual features: the visible light image is a high-resolution color image with rich textures and details, while the infrared image is monochrome, low contrast, and lack textures. Besides, because of the different clock frequencies of the different photosensitive chips, even if the two cameras are registered in advance, the position of the same object in the two images is not necessarily the same. Unregistered cases require that the fusion tracker have a certain ability of misalignment prevention during the fusion process.

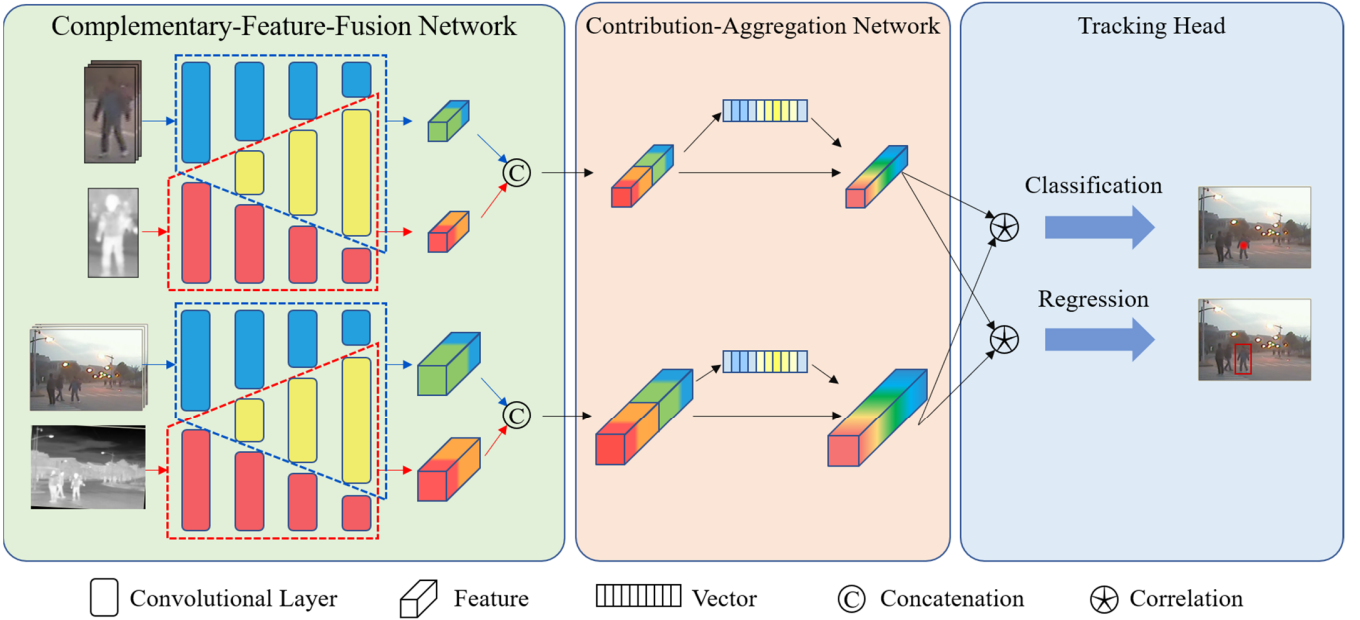

Fig. 2. Illustration of the proposed SiamIVFN framework. The complementary-feature-fusion network (CFFN) is used to extract and fuse the features of RGB-T image pairs. The contribution-aggregation network (CAN) is utilized to calculate the contribution of different features for tracking tasks adaptively. The tracking head is divided into two branches: classification and regression. Please refer to Section 3 for more details.

Based on the above analysis, a feature-level fusion network, called Siamese Infrared and Visible Light Fusion Network (SiamIVFN), is proposed to track an object in RGB-T image pairs. The feature fusion part of SiamIVFN is composed of two subnetworks: a complementary-feature-fusion network (CFFN) and a contribution-aggregation network (CAN). CFFN uses a two-stream convolution structure to extract and fuse the features of infrared and visible light images. In each layer of the two-stream convolution, a coupled filter is designed to extract the common features from the image pairs. Considering that the similarity of the features extracted from the shallow layers in the RGB-T image pairs is different from the features extracted from the deep layers, we gradually increase the coupling rates. The effectiveness of the coupling rate setting is demonstrated in the experimental part.

Besides, because infrared and visible light images have different contributions to object tracking, further processing of the feature fusion should be considered. CAN uses the self-attention method to adaptively calculate the contribution of infrared and visible light images to different visual conditions. Experiments show that SiamIVFN achieves the best effects of infrared and visible light fusion tracking.

In general, SiamIVFN has three main contributions:
● A complementary-feature-fusion subnetwork (CFFN), a two-stream convolutional network with coupling filters, is proposed to extract the common and individual features between infrared and visible light image pairs.
● A contribution-aggregation subnetwork (CAN) is designed to make the fusion tracker adaptively calculates the contributions of the infrared and visible light features obtained from CFFN.
● The proposed SiamIVFN is a siamese-framework-based fusion tracker for RGB-T fusion tracking. Real-time tracking is always considered in the design of SiamIVFN, so the structure of SiamIVFN is straightforward. Experimental results show that the tracking speed of SiamIVFN is fast (147.6FPS).

## II. RELATED WORK

### A. Visual Object Tracking

In object tracking, deep learning-based trackers have achieved state-of-the-art performance on multiple public datasets with their powerful representation capabilities. At present, most object trackers based on deep learning have adopted the structure of the siamese network. SiamFC [2] used the similarity learning method to treat the object tracking problem as a template matching problem. SiamFC is simple and fast. However, since SiamFC uses a multi-scale prediction method, it cannot handle the situation when the size of the object changes drastically. To solve this problem, SiamRPN [18] introduced the region proposal network (RPN). Researchers have improved the siamese-network-based methods in data preprocessing [19], network structure [29], and multilayer feature fusion [22]. SiamFC++ [34] introduced the concept of anchor-free in the siamese network, thus improving the speed and ac-curacy of the siamese-framework-based tracker. Although visible light object tracking can achieve good results, it still cannot handle smoke, night, and other bad visual conditions due to the characteristic of RGB sensors.

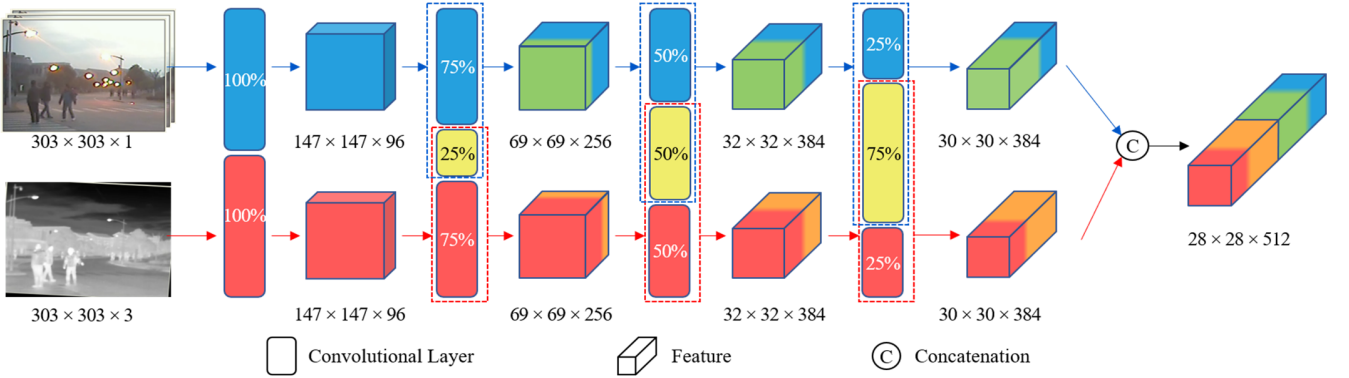

Fig. 3. Illustration of CFFN. CFFN is a two-stream convolutional network, which extracts the features of infrared and visible light images, respectively. The two-stream convolutional network is equipped with coupling filters with different coupling rates.

*B. RGB-T Object Tracking*

Owing to the penetration ability of infrared sensors, they are often adopted to work with RGB sensors. Therefore, RGB-T fusion tracking has recently attracted more and more attention. SiamFT [27] and DsiamFT [38] used siamese networks to solve the RGB-T fusion tracking problem. They used different backbones to extract features from infrared and visible light image pairs then merged them and fed them into the tracking head. Due to the simple structure of the siamese-base methods, the tracking speed of these methods is fast. However, these fusion methods processed the image pairs separately. They did not fully consider the common features of infrared and visible light images, resulting in a lot of feature redundancy and computational burden. Unlike extracting features using different backbones, MANet [24] and CANet [31] shared a part of the same convolution kernels (or so-called coupling filters) to extract common features of infrared and visible light images. However, in designing convolution kernels of different depths, they did not fully consider the features extracted by different depth layers. In this paper, we use different coupling filters in different layers. Besides the consideration of the common feature extraction, the attention mechanism is utilized to extract individual features that reflect the characteristics of the two different sensors. The experimental results of LTDA [25] and CMPP [33] showed that the attention mechanism could largely improve tracking performances.

## III. OUR METHOD

This section will introduce the proposed SiamIVFN. First, we summarize the overall structure of SiamIVFN and then introduce the structure of CFFN and CAN.

*A. The Architecture*

The SiamIVFN network consists of three parts: CFFN, CAN, and tracking head. The structure of SiamIVFN is illustrated in Fig. 2. In the online tracking process, given the infrared and visible light video sequences, the tracker will track the position of the object in each frame. Compared with visible light images, infrared images lack certain specific information, such as color and texture. It is necessary to use uncoupled filters to extract individual features from infrared and visible light images separately. Since each infrared image and visible image pair simultaneously capture the same scene, they contain common features, such as semantics and contours. Wang [8] argued that coupled filters could extract common features. Li [32] adopted the coupled filters for depth estimation and showed their effectiveness. Inspired by this research works, this paper proposes a complementary-feature-fusion network (CFFN) to extract and fuse the features of infrared and visible light images.

Besides the common features, infrared and visible light images contain individual features for object tracking and may have different contributions to tracking tasks under different light conditions. In degraded light conditions such as fog and night, infrared images contribute more than visible light images for object tracking. While under normal lighting conditions, visible light images are more suitable than infrared images to detect and track an object. Most existing fusion methods regard the contribution of infrared image and visible image as the same and often directly concatenate the features extracted from infrared and visible light images. In this paper, the contribution-aggregation network (CAN) is proposed, which adaptively calculates the contribution of different features. CAN utilizes the self-attention module [16] to adaptively calculate the contributions of infrared and visible light images according to different light conditions.

*B. Complementary-Feature-Fusion Network*

The details of the CFFN are depicted in Fig. 3. CFFN adopts a two-stream convolution structure. The lower branch represents the convolutional flow of infrared images. The upper branch represents the convolutional flow of visible light images. Unlike other two-stream networks, CFFN sets up filters with different coupling rates in each convolutional layer to learn the common features between infrared and visible images. The overlapping red part between the two indicates the coupling part of the two image filters. In this way, infrared and visible light images are mutually auxiliary. The features extracted from the infrared image are supplementary to the stream network designed for the visible light image. In the other stream network for infrared images, features extracted from visible light images are supplementary through partially coupled filters.

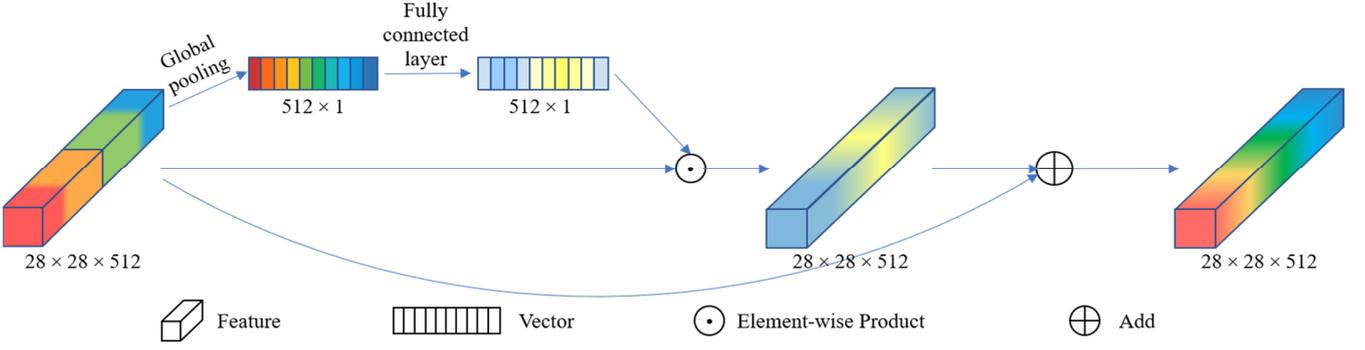

Fig. 4. Illustration of CAN. CAN first reform the extracted features into channel-wise vectors and then passes through two fully connected layers. The output of the second connected layer is then multiplied with the original feature. The multiplied feature is then added with the original feature.

The uncoupled filters are designed to learn the individual features. The ratio of the number of coupling filters to the number of all filters is called the coupling ratio:

$$R_i = \frac{k_i}{n_i} \ (i = 1, 2, 3, 4) \quad (1)$$

where $R_i$ is the coupling rate of the $i$th layer, $k_i$ is the number of coupled filters in the $i$th convolutional layer, and $n_i$ is the number of all filters in the $i$th convolutional layer. We set the coupling ratio of each convolutional layer: 0.25, 0.5, and 0.75. In Section 4, we use the grid search method to demonstrate the effectiveness of the coupling rate. The coupling rate increases as the convolutional layers go deeper.

In CFFN, the parameters of the filter are updated through the backpropagation algorithm. In each iteration, the non-coupling filter for infrared and visible light images is updated once, and the coupling filter for infrared and visible light images is updated twice:

$$w_{RGB}^{(i)} = \begin{cases} w_{RGB_{ucoupled}}^{(i-1)} + l\frac{\partial L}{\partial w_{RGB_{ucoupled}}^{(i-1)}} \\ w_{coupled}^{(2i-1)} + l\frac{\partial L}{\partial w_{coupled}^{(2i-1)}} \end{cases} \quad (2)$$

$$w_T^{(i)} = \begin{cases} w_{T_{ucoupled}}^{(i-1)} + l\frac{\partial L}{\partial w_{T_{ucoupled}}^{(i-1)}} \\ w_{coupled}^{(2i-2)} + l\frac{\partial L}{\partial w_{coupled}^{(2i-2)}} \end{cases} \quad (3)$$

where $w$ is the parameter that needs to be updated, $(i)$ is the number of iterations, $l$ is the learning rate, and $L$ is the loss function.

In summary, a two-stream convolutional structure is designed in CFFN. Besides individual features, the two-stream convolutional blocks are able to extract common features by the coupling filters.

### C. 3.3. Contribution-Aggregation Network

After extracting features from infrared and visible light images (using certainly separated backbones), most of the existing fusion trackers directly concatenate the features and then send them to the tracking head for tracking. However, because the different features have different contributions to object tracking, especially under various light conditions, this paper proposes CAN to adaptively calculate the contribution of the features. The details of CAN are shown in Fig. 4.

CAN first utilizes global average pooling to each channel to obtain a global feature $g_c$ as:

$$g_c = \frac{1}{H \times W} \sum_{i=1}^{H} \sum_{j=1}^{W} x_c(i, j) \quad (4)$$

The global feature then passes through two fully connected layers to improve the generalization ability of CAN:

$$h_c = \beta(\alpha(g_c)) \quad (5)$$

Where $\alpha(\cdot)$ and $\beta(\cdot)$ are two different fully connected layers. The learned feature vector $h_c$ is multiplied by the original feature $x_c$:

$$y_c = h_c \cdot x_c \quad (6)$$

Finally, the obtained feature $y_c$ is added to the original feature to calculate the output of CAN:

$$z_c = y_c + x_c \quad (7)$$

The whole procedure of CAN could be viewed as learning the weight coefficient of each channel through self-attention, which pays more attention to the channels critical for object tracking through end-to-end learning.

## IV. IMPLEMENTATION DETAILS

This section will introduce the training and online tracking process. The tracking head is build based on SiamFC++ [34]. We train and test SiamIVFN on the PyTorch platform with Intel® Core™ I7-10700K CPU and NVIDIA® TITAN RTX™ GPU.

### A. Training Procedure

*1) Pre-training.*

We use the GOT10K [17] and LASOT [20] datasets to train our network end-to-end. Since GOT10K and LASOT are both RGB datasets, they do not have infrared images. We use visible light images to generate grayscale images to train coupling filters and non-coupling filters. The optimization algorithm is the stochastic gradient descent method with momentum. The momentum is set to 0.9, and the weight attenuation is set to 0.0001. The learning rate adopts the cosine decay strategy, the initial learning rate is set to 0.08, and the final learning rate is set to 1e-6.

*2) Training.*

Based on the pre-trained network, we further train the entire network by the RGB-T dataset. In the first ten epochs, CFFN is fixed to train CAN and the tracking head. In the second ten epochs, we unfreeze the non-couping filters for infrared images in CFFN. In the third ten epochs, we unfreeze the coupling filter in CFFN. After the 40th period, we unfreeze the whole CFFN for training. Such gradual training can accelerate the convergence of the network. To improve the discriminative ability of the network, we set the maximum index of a pair of sample frames to 1000 and the ratio of the number of positive sample pairs to the number of negative sample pairs to 0.5. In terms of optimization algorithms, we use Adam to optimize the loss function. The learning rate also uses cosine decay. The initial learning rate is set to 8e-5, and the end learning rate is set to 1e-6.

TABLE I
RGBT234 DATASET PR/SR SCORES BASED ON ATTRIBUTES. THE BEST, SECOND-BEST, AND THIRD-BEST PR (SR) ARE SHOWN IN RED, BLUE, AND YELLOW.

| PR/SR | KCF | ECO | C-COT | MDNet | SiameseFC | MANet | MACNet | SGT | DAPNet | DAFNet | SiamIVFN |
|---|---|---|---|---|---|---|---|---|---|---|---|
| NO | 57.1/37.1 | 88.0/65.5 | 88.8/65.6 | 81.2/59.0 | 76.6/55.5 | 88.7/64.1 | 75.4/50.4 | 87.7/55.5 | 90.0/64.4 | 90.0/63.6 | 86.0/68.5 |
| PO | 52.6/34.4 | 72.2/53.4 | 74.1/54.1 | 74.7/50.9 | 62.8/45.0 | 81.7/56.2 | 70.1/46.5 | 77.9/51.3 | 82.1/57.4 | 85.9/58.8 | 83.0/65.3 |
| HO | 35.6/23.9 | 60.4/43.2 | 60.9/42.7 | 63.3/43.2 | 53.9/36.9 | 68.8/46.1 | 53.2/34.6 | 59.2/39.4 | 66.0/45.7 | 68.6/45.9 | 77.3/58.9 |
| LI | 51.8/34.0 | 63.5/45.0 | 64.8/45.4 | 58.9/39.6 | 58.5/40.2 | 76.9/51.3 | 69.7/45.5 | 70.5/46.2 | 77.5/53.0 | 81.2/54.2 | 81.0/62.1 |
| LR | 49.2/31.3 | 68.7/46.4 | 73.1/49.4 | 66.0/44.5 | 62.6/41.4 | 76.0/51.1 | 57.8/35.1 | 75.1/47.6 | 75.0/51.0 | 81.8/53.8 | 73.9/55.4 |
| TC | 38.7/25.0 | 82.1/60.9 | 84.0/61.0 | 74.8/53.0 | 61.2/43.2 | 75.4/53.8 | 48.4/31.9 | 76.0/47.0 | 76.8/54.3 | 81.1/58.3 | 64.9/48.3 |
| DEF | 41.0/29.6 | 62.2/45.8 | 63.2/46.3 | 66.4/46.8 | 58.1/42.6 | 72.0/52.0 | 61.6/42.5 | 68.5/47.4 | 71.7/51.8 | 74.1/51.6 | 79.6/63.0 |
| FM | 37.9/22.3 | 57.0/39.5 | 62.8/41.8 | 63.2/39.3 | 57.5/37.6 | 69.5/44.5 | 55.7/33.3 | 67.7/40.2 | 67.0/44.3 | 74.0/46.5 | 67.0/48.3 |
| SV | 44.1/28.7 | 74.0/55.8 | 76.2/56.2 | 73.9/51.9 | 63.0/45.2 | 77.7/53.9 | 65.0/42.8 | 69.2/43.4 | 78.0/54.2 | 79.1/54.4 | 82.9/65.3 |
| MB | 32.3/22.1 | 68.9/52.3 | 67.3/49.5 | 62.4/44.2 | 56.4/40.8 | 72.6/51.3 | 46.8/32.2 | 64.7/43.6 | 65.3/46.7 | 70.8/50.0 | 63.3/49.7 |
| CM | 40.1/27.8 | 63.9/47.7 | 65.9/47.3 | 61.3/43.3 | 58.0/41.8 | 71.9/50.3 | 55.5/37.9 | 66.7/45.2 | 66.8/47.4 | 72.3/50.6 | 70.6/54.7 |
| BC | 42.9/27.5 | 57.9/39.9 | 59.1/39.9 | 62.5/41.8 | 50.2/33.7 | 73.8/48.0 | 57.3/34.7 | 65.8/41.8 | 71.7/48.4 | 79.1/49.3 | 76.9/58.0 |
| ALL | 46.3/30.5 | 70.2/51.4 | 71.4/51.4 | 71.0/49.0 | 61.7/43.6 | 77.7/53.5 | 63.9/42.2 | 72.0/47.2 | 76.6/53.7 | 79.6/54.4 | 81.1/63.2 |

### B. Online Tracking

In the online tracking process, the template RGB-T image pair and the RGB-T image pair to be searched feed to CFFN. Then CAN obtain the features of the template and the search area. After the two features are cross-correlated, a score map is computed for classification. According to Xu [34], the direct utilization of the score map for boundary selection might cause performance degradation. In this paper, we adopt a quality estimation branch in addition to the classification branch. The classification branch uses focal loss [39]. The quality estimation branch uses center loss [40]. The $1 \times 1$ convolution is used to weight the classification score and quality estimation score to get the overall classification score. In the regression branch, to avoid artificially setting anchor points and thresholds and other manual intervention, we adopt the idea of anchorless to directly predict the four sides from the corresponding position $(x, y)$ in the bounding box. The regression branch uses IOU loss [34].

## V. EXPERIMENT

### A. Evaluation Dataset and Evaluation Metrics

We compare SiamIVFN and other tracking methods on two RGB-T tracking benchmark datasets to demonstrate the performances. The GTOT [5] dataset has 15.8K frames, containing 50 RGB-T videos aligned in space and time and seven annotated attributes. The RGBT234 [23] dataset has 234K frames, 234 aligned RGB-T videos and twelve annotated attributes. Due to the large differences in the number, quality, and data distribution of GTOT and RGBT234, we divided GTOT and RGBT234 into different training sets and test sets, respectively. We divide GTOT into five parts, each containing ten videos. When performing experiments on GTOT, we use four parts for training and one part for testing. We conduct five separate experiments to ensure that all GTOT datasets are tested. When performing experiments on RGBT234, we divide the dataset into nine parts, each containing 26 videos. Eight parts are utilized for training, while one left part for testing. Nine experiments were performed separately.

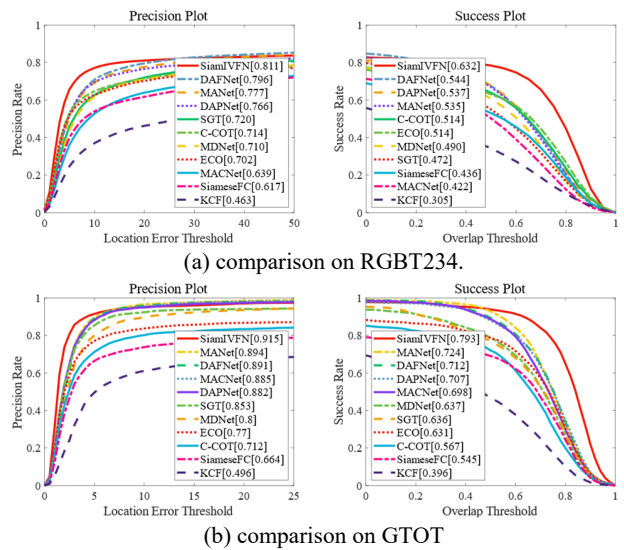

Fig. 5. Overall performance compared with state-of-the-art trackers on RGBT234 (a) and GTOT (b).

The precision rate (PR) and the success rate (SR) in one-pass evaluation (OPE) are used as evaluation indicators. PR refers to the percentage of frames whose distance between the output position and the ground truth position is within a threshold. We set the thresholds of GTOT and RGBT234 datasets to 5 and 20, respectively. SR is the proportion of frames whose overlap ratio between the output bounding box, and the ground truth bounding box is larger than a threshold. We use the area under the curves (AUC) to calculate the SR score.

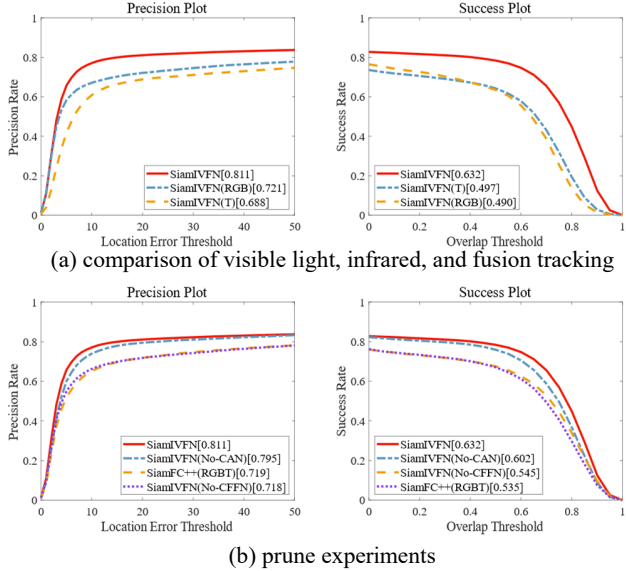

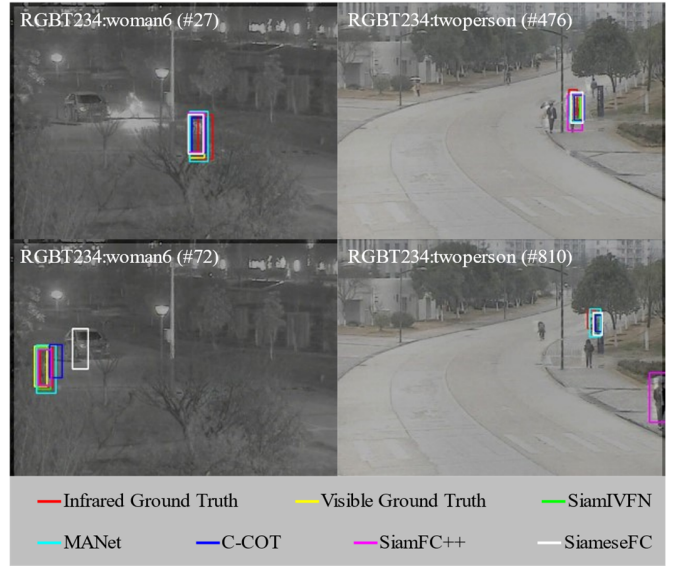

Fig. 6. Comparison of visible light, infrared, and fusion tracking (a). Ablation experiment (b) on RGBT234.

Fig. 7. Qualitative analysis of SiamIVFN and other trackers.

### B. Comparison with State-of-the-art Trackers

We implemented SiamIVFN on the GTOT and RGBT234 benchmarks and compared them with other state-of-the-art RGB trackers (KCF [1], ECO [10], C-COT [3], MDNet [7], and SiameseFC [2]) and state-of-the-art fusion trackers (SGT [12], MANet [24], MACNet [35], DAPNet [30], and DAFNet [21]). The overall tracking performances are shown in Fig. 5. It can be found that SiamIVFN outperforms other trackers. Specifically, on the RGBT234 dataset, the PR/SR score of SiamIVFN reached 81.1%/63.2%, 1.5%/8.8% higher than the second-best method. As for the GTOT dataset, the PR/SR score of SiamIVFN reached 91.5%/79.3%, 2.1%/6.9% higher than the second-best one. The experimental results demonstrated the effectiveness of the proposed SiamIVFN.

To further show the performances of SiamIVFN, we separately calculated the PR/SR scores of each attribute in the RGBT234 dataset. The specific results are recorded in Table I. It can be concluded from the table that SiamIVFN has the highest scores in almost all attributes than other trackers.

Besides the improvement of precision rates, the success rates of the proposed SiamIVFN is much higher than any other tracker (8.8% higher than the second-best), specifically under the challenges of low illumination (LI), low resolution (LR), background clutter (BC), partial occlusion (PO), and heavy occlusion (HO). It means that the proposed subnetworks CFFN and CAN are able to adaptively extract and fuse the features for the success of object tracking. The two-stream convolutional structure and the channel-wise aggregation are simple and effective for the RGB-T tracking tasks.

In the case of low illumination (LI), relying only on visible light for tracking will lead to poor results. Since SiamIVFN can integrate visible light and infrared images and use infrared image information to supplement tracking, the success rate of SiamIVFN increases by 7.9% compared to the second-best (DAFNet). In the case of background clutter (BC), because of the simple background of the infrared images, SiamIVFN exploits the individual features of the infrared images, and the success rate of SiamIVFN is 8.7% higher than the second-best (DAFNet). In the case of partial occlusion (PO) and heavy occlusion (HO), SiamIVFN can extract the common features and cope with a certain misalignment, thereby increasing the robustness of tracking. In PO and HO, the success rate of SiamIVFN increases by 6.5% and 12.8% from the second-best (DAFNet for PO, MANet for HO).

### C. Ablation Study

In this subsection, we compared the tracking performances of SiamIVFN (RGB), SiamIVFN (T), and SiamIVFN. SiamIVFN (RGB) and SiamIVFN (T) respectively indicate that SiamIVFN relies solely on visible light and infrared images for tracking. SiamIVFN (RGB) refers to replacing the infrared image part with the visible light image. SiamIVFN (T) refers to replacing the visible light image part with the infrared image. The tracking performance is shown in Fig. 6(a). The experimental results show that SiamIVFN fusion tracking is significantly better than relying solely on infrared images (12.3%/12.6) or visible light images for tracking (9.0%/14.2%).

To show the performance of the two subnetworks, CFFN and CAN, we remove CFFN and CAN from SiamIVFN (denoted by SiamIVFN (No-CFFN) and SiamIVFN (No-CAN)). Comparative experiments are performed on the RGBT234 dataset. SiamFC++(RGBT) is the baseline. In SiamFC++(RGBT), the infrared image is directly used as the fourth channel, concatenated on the RGB image, and then tracked by SiamFC++. The results are shown in Figure 6(b), which show that:

1) Comparing the performance of SiamIVFN and SiamIVFN (No-CAN), the PR/SR score with CAN improves by 1.6%/3.0%.

2) Comparing the performance of SiamIVFN and SiamIVFN (No-CFFN), the PR/SR score with CFFN improves by

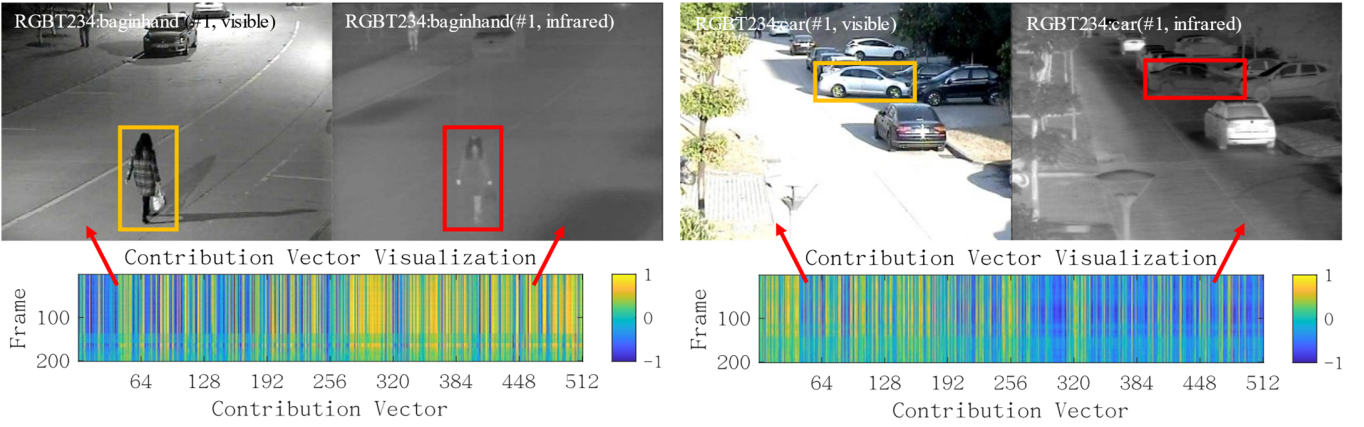

Fig. 8. Visualization of contribution vectors of CAN. The horizontal axis represents 512 vectors, and the vertical axis represents the number of frames in the video sequence. Color from cold to warm represents the value from 0 to 1.

TABLE II
COMPARISON OF DIFFERENT COUPLING RATES ON RGBT234.

| Experiment (#) | Conv2 | Conv3 | Conv4 | Presion Rate | Success Rate |
|---|---|---|---|---|---|
| 1 | 0.25 | 0.50 | 0.75 | 81.1 | 63.2 |
| 2 | 0.25 | 0.75 | 0.50 | 72.9 | 52.3 |
| 3 | 0.50 | 0.25 | 0.75 | 72.5 | 54.3 |
| 4 | 0.50 | 0.75 | 0.25 | 66.5 | 47.6 |
| 5 | 0.75 | 0.25 | 0.50 | 70.9 | 50.1 |
| 6 | 0.75 | 0.50 | 0.25 | 64.0 | 45.1 |

9.3%/8.7%.

The coupling rate of different layers in the CFFN is a hyperparameter of SiamIVFN. We arrange the coupling rates {0.25, 0.5, 0.75} in separate layers and then compare networks with different coupling rates. The tracking performance under RGBT234 is shown in Table II. When the coupling ratio is {0.25, 0.5, 0.75}, the tracker can obtain the best performance. The features extracted by the shallow layers are individual features such as color and texture. These individual features between infrared and visible light images are quite different, so the appropriate coupling rate is small. On the contrary, the common features such as the contour extracted by the deep network, between infrared and visible light images, are relatively similar, so the appropriate coupling rate is larger.

### D. Qualitative performances

To visually show the tracking performances of SiamIVFN, we took four sequences for comparison. Fig. 7 shows the bounding box of SiamIVFN and that of other trackers (MANet, C-COT, SiamFC, and SiamFC++). To show the bounding boxes in one image, we linearly superimpose the infrared images and the visible light images. Red boxes and yellow boxes are utilized to frame the ground truth position of the target originally given in the infrared images and the visible light images.

The two images of the second column are selected from woman6, whose background is complex. The complex background can easily interfere with the classification score of the object, making it impossible to distinguish the foreground and background correctly. The infrared image background is simple and easy to distinguish. The CFFN extracts the individual features of infrared and visible images, improving the stability of the tracker through the infrared port in complex backgrounds.

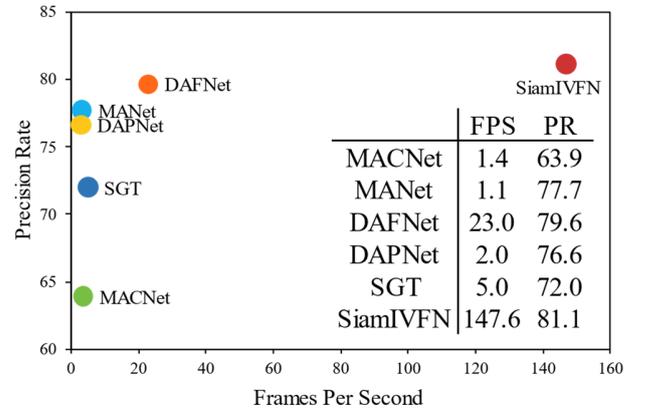

(a) Comparison of precision rate and speed

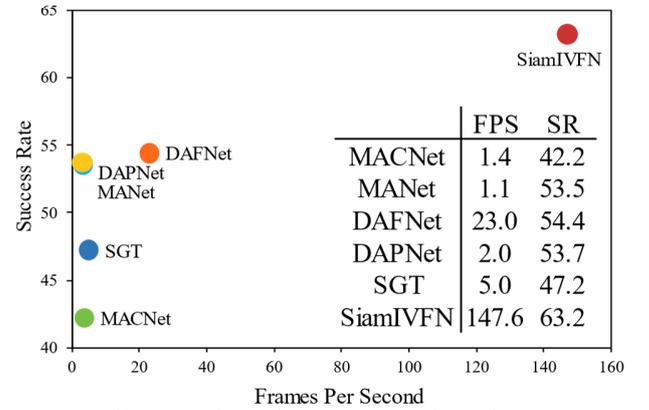

(b) Comparison of success rate and speed

Fig. 9. Speed comparison of various tracking methods

The false object was occluding the real object in twoperson, the third column. When the fake object pass by the real object and misalignment occurred, the infrared part of the real object is located in the visible part of the fake object, leading to the classification score of the fake object is higher than the real object, causing the tracker to make an error. Since the CFFN is a feature-level fusion, it can cope with slight misalignment.

To illustrate the effectiveness of the CAN, 200 frames are selected from daytime sequences and nighttime sequences of the RGBT dataset, respectively. We visualize the contribution vectors for some frames in Fig. 8. The first 256 contribution vectors in the figure are calculated from visible light features, and the 257th to 512th contribution vectors are calculated from infrared features. It can be found that in the nighttime sequences *beginhand*, the infrared feature has larger contribution weights (warm color). In contrast, in the daytime sequence *car*, the contribution weights of the visible light feature are relatively large. It means that the CAN pay more attention to features beneficial to the tracking task.

*E. Efficiency Analysis*

We compare the efficiency of SiamIVFN with that of other fusion tracking methods in Fig. 9. It can be found that the speed of the proposed SiamIVFN greatly exceeds other fusion methods. SiamIVFN reached 147.6FPS, which is 124.6FPS faster than the second-best fusion tracker DAFNet. In the design of SiamIVFN, we give priority to speed take the siamese-based structure as the tracking head. Besides, both CFFN and CAN are more concise than the backbone of other fusion tracking methods.

## VI. CONCLUSION

A novel RGB-T image-based tracking method, called SiamIVFN, is proposed in this paper. SiamIVFN can fuse the complementarity of infrared and visible light images adaptively to address the object tracking problem under various light conditions. SiamIVFN mainly contains two subnetworks, CFFN and CAN.

Owing to the two-stream convolutional structure, CFFN can extract both common features and individual features from infrared and visible light image pairs. CFFN treats infrared and visible light images as complements for each other through the coupling filters. The common features of infrared and visible light images can be learned without additional computation. CFFN is a feature-level fusion network in nature, which is able to handle the situation where visible light images and infrared images are not rigorous aligned. Under various light conditions, CAN is designed to adaptively compute the contributions of different features, which could learn the weight coefficient of each channel through self-attention. Experiments performed on two RGB-T tracking benchmark datasets demonstrate that SiamIVFN outperforms other latest RGB-T tracking methods and can reach 147.6FPS.